\documentclass[letterpaper, 10 pt, conference]{ieeeconf}  

\IEEEoverridecommandlockouts                              

\usepackage{graphics} 
\usepackage{amsmath} 
\usepackage{amssymb}  
\usepackage{hyperref}
\usepackage{multirow}
\usepackage{booktabs}
\usepackage{makecell}
\usepackage{cuted}
\usepackage{subcaption}
\usepackage{stfloats} 
\usepackage{tabularx}
\usepackage{graphicx}
\usepackage{float}

\usepackage{colortbl}
\usepackage[dvipsnames]{xcolor}
\usepackage{caption}
\usepackage{placeins}

\usepackage{marvosym}
\usepackage{pifont}
\usepackage{stfloats}

\usepackage[utf8]{inputenc} 
\usepackage[utf8]{inputenc} 
\usepackage{newunicodechar}
\newunicodechar{，}{,}
\usepackage[T1]{fontenc}    
\usepackage{pifont}
\usepackage{url}            
\usepackage{booktabs}       
\usepackage{amsfonts}       
\usepackage{nicefrac}       
\usepackage{microtype}      
\usepackage{xcolor}         
\usepackage{listings}
\usepackage{graphicx}
\usepackage{array}
\usepackage{amsmath}
\usepackage{amssymb}
\usepackage{adjustbox}
\usepackage{tcolorbox}
\usepackage{multirow}
\usepackage{makecell} 
\usepackage{tabularx}

\usepackage{booktabs}
\usepackage{multicol}
\usepackage{array}
\usepackage{xcolor}
\usepackage{colortbl}

\title{\LARGE \bf
FieldGen: From Teleoperated Pre-Manipulation Trajectories to Field-Guided Data Generation}

\author{Wenhao Wang$^{*2}$, Kehe Ye$^{*2,4}$, Xinyu Zhou$^{*2,6}$, Tianxing Chen$^{*3,4}$, Cao Min$^{6}$, Qiaoming Zhu$^{6}$, \\Xiaokang Yang$^{1}$, Ping Luo$^{3}$, Yongjian Shen$^{2}$, Yang Yang$^{2}$, Maoqing Yao$^{2}$, Yao Mu$^{\text{\Letter} 1,5}$\\\\
$^{1}$MoE Key Lab of Artificial Intelligence, AI Institute, SJTU, $^{2}$AgiBot, $^{3}$HKU, $^{4}$Lumina Group, \\
$^{5}$Shanghai AI Laboratory, 
$^{6}$School of Computer Science and Technology, Soochow University\\
$\textsuperscript{*}$ Equal contribution \quad $^{\text{\Letter}}$ Corresponding authors \quad Wegpage: \href{https://fieldgen.github.io/}{https://fieldgen.github.io/}
\vspace{-10pt}
}

\begin{document}

\vspace{-20pt}
\twocolumn[{%
  \renewcommand\twocolumn[1][]{#1}%
  \maketitle
  \vspace{-3pt}
  \begin{center}
    \includegraphics[width=\textwidth]{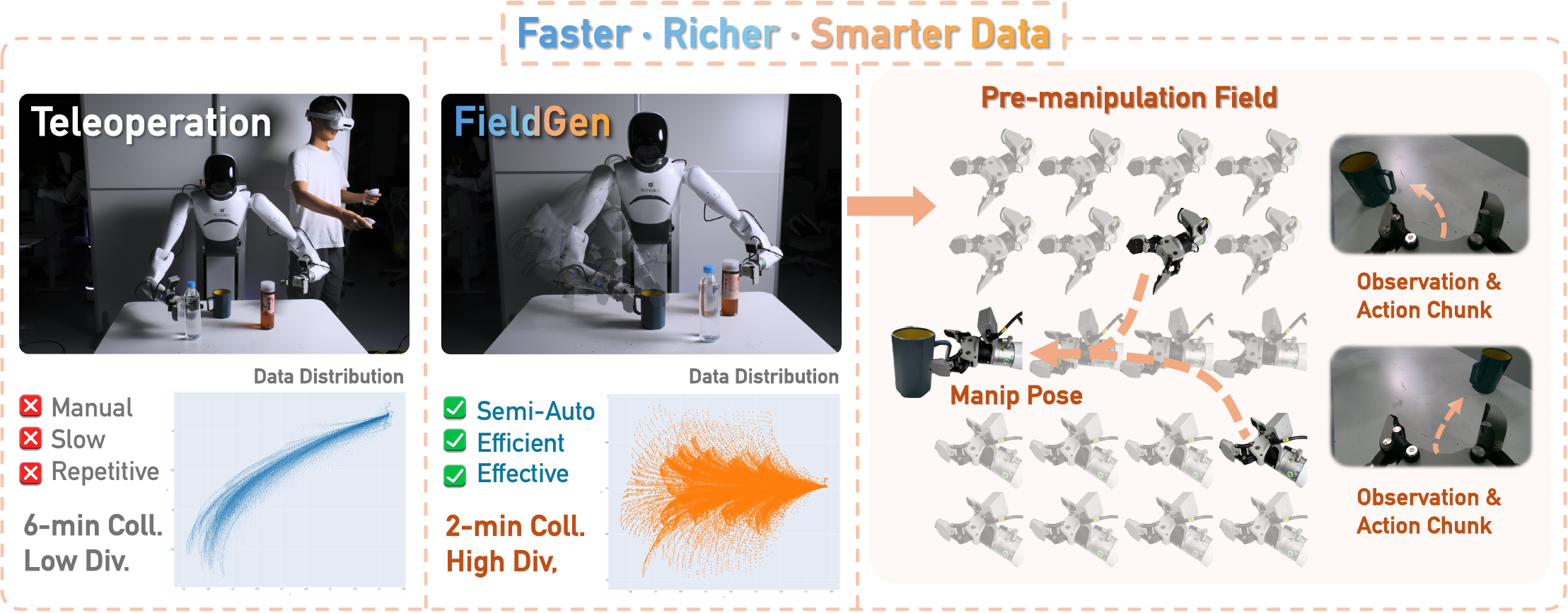}
    \vspace{-10pt}
    \captionsetup{type=figure}
    \caption{\textbf{FieldGen} is a semi-auto data generation framework that enables scalable collection of diverse, high-quality real-world manipulation data with minimal human involvement.}
    \label{fig:teaser}
  \end{center}
}]

\thispagestyle{empty}
\pagestyle{empty}

\begin{abstract}

Large-scale and diverse datasets are vital for training robust robotic manipulation policies, yet existing data collection methods struggle to balance scale, diversity, and quality. Simulation offers scalability but suffers from sim-to-real gaps, while teleoperation yields high-quality demonstrations with limited diversity and high labor cost.
We introduce FieldGen, a field-guided data generation framework that enables scalable, diverse, and high-quality real-world data collection with minimal human supervision. FieldGen decomposes manipulation into two stages: a pre-manipulation phase, allowing trajectory diversity, and a fine manipulation phase requiring expert precision. Human demonstrations capture key contact and pose information, after which an attraction field automatically generates diverse trajectories converging to successful configurations.
This decoupled design combines scalable trajectory diversity with precise supervision. Moreover, FieldGen-Reward augments generated data with reward annotations to further enhance policy learning. Experiments demonstrate that policies trained with FieldGen achieve higher success rates and improved stability compared to teleoperation-based baselines, while significantly reducing human effort in long-term real-world data collection. Webpage is available at \href{https://fieldgen.github.io/}{https://fieldgen.github.io/}.

\end{abstract}
\section{INTRODUCTION}
Recent end-to-end embodied intelligence models have achieved promising progress in robotic manipulation, demonstrating strong generalization capabilities and few-shot performance~\cite{black2024pi0visionlanguageactionflowmodel,intelligence2025pi05visionlanguageactionmodelopenworld,liu2024rdt,chen2025benchmarking,chen2025g3flow,lu2024manicm}. However, these models critically depend on large-scale, diverse datasets to achieve effective performance. The collection of real-world robotic data requires substantial human effort through teleoperation and manual demonstration~\cite{bu2025agibot, vuong2023open,liu2025avr}, making the creation of large-scale datasets prohibitively expensive.

Existing data collection approaches face a fundamental trade-off between scale, diversity, and data quality. Simulation-based methods can generate vast amounts of data with spatial randomization, but suffer from persistent sim-to-real gaps and limited behavioral diversity~\cite{mu2024robotwin, chen2025robotwin, jiang2025dexmimicgen, mandlekar2023mimicgen}. Conversely, teleoperation produces high-quality demonstrations but faces severe scalability constraints: operators control only one robot at a time, experience cognitive fatigue during extended sessions, and converge to stereotypical motion patterns despite explicit instructions for diversity. Our analysis of large-scale teleoperation datasets reveals pronounced multimodal distributions that burden policy learning compared to more uniform behavioral diversity. Semi-automated approaches like PATO~\cite{dass2022pato} and GCENT~\cite{wang2025genie} attempt to bridge this gap but still rely on pre-trained policies and may not escape human behavioral constraints.

To address these challenges, we present \textbf{FieldGen}, a field-guided data generation framework that enables scalable collection of diverse, high-quality real-world manipulation data with minimal human effort. Our key observation is that manipulation naturally decomposes into two phases with distinct requirements: a pre-manipulation phase of reaching and approaching, where trajectory variation is acceptable as long as paths converge to valid manipulation configurations, and a fine-manipulation phase requiring precise, contact-rich control, for which expert demonstrations are most valuable. FieldGen implements a semi-automated pipeline: we first gather a small set of high-quality human demonstrations emphasizing critical poses and contact interactions. From these, we construct an attraction field that guides pre-manipulation trajectories toward the identified manipulation configurations. Automated scripts then sample randomized initial observations and compute trajectories that asymptotically converge to this field, yielding large volumes of observation–action pairs for the pre-manipulation phase. This decoupling provides three benefits: (i) high-quality supervision for fine manipulation via focused human demos, (ii) scalable automated generation of diverse pre-manipulation trajectories, and (iii) efficient, large-scale data collection that preserves both diversity and quality while greatly reducing human involvement relative to teleoperation. We further introduce FieldGen-Reward, which generates trajectories of varying quality with explicit labels, enabling models to better capture behavioral causality and improving the robustness of learned manipulation policies.

We summarize our contributions as follows: (1) \textbf{Pre-manipulation field (PMF)}. We introduce the pre-manipulation field as a new concept, defined as an abstract representation that models and extrapolates from a small set of demonstration trajectories. PMF provides a unified framework for large-scale, semi-automated, real-robot data collection. (2) \textbf{Semi-automated collection via phase decoupling}. Building on PMF, we introduce a semi-automated pipeline that decouples manipulation into pre-manipulation and fine-manipulation phases: teleoperated demonstrations provide critical manipulation poses, from which we synthesize large-scale datasets for the pre-manipulation phase. The pipeline further produces trajectories of varying quality with explicit quality annotations, substantially improving automated collection efficiency while preserving diversity and fidelity. (3) \textbf{Empirical validation}. Experiments show that our method produces larger, higher-quality datasets in less time; models trained on our data are more robust; and operator workload is markedly reduced, enabling long-duration, stable data collection.

\section{RELATED WORKS}

\subsection{Teleoperation for Robotic Data Collection}
Teleoperation with VR and custom hardware has been widely adopted to collect high-fidelity real-robot demonstrations. Systems leverage head-mounted displays~\cite{cheng2024open,ze2024humanoid_manipulation}, cameras~\cite{qin2023anyteleop,handa2019dexpilot}, hand controllers or gloves~\cite{dass2024telemoma,2405f1ce7e494be0a83d98072e757122,wang2024dexcap}, and low-cost bimanual platforms~\cite{wu2023gello,fang2023airexo,fang2025airexo2,ben2025homie,fu2024mobilealoha,li2025simplevla} to improve precision, reduce latency, and raise operator throughput. However, teleoperation suffers from fundamental scalability and quality limitations as human operators experience cognitive fatigue and unconsciously converge to stereotypical motion patterns, resulting in expensive data collection with insufficient diversity for training robust robotic policies. Recent efforts have attempted to improve teleoperation efficiency through assistance mechanisms. Recent efforts have attempted to address these limitations through assistance mechanisms. PATO~\cite{dass2022pato} employs learned policies to automate repetitive subtasks while preserving human oversight for critical decisions, effectively reducing operator cognitive load, and GCENT~\cite{wang2025genie} expanded it to real-world long-horizon tasks. However, such approaches still rely on pre-trained policies and may inherit the diversity limitations inherent in human demonstrations, as they automate existing human-demonstrated subtasks rather than fundamentally changing the data generation paradigm. In contrast, our approach differs by decomposing manipulation tasks at a more fundamental level and automatically generating diverse pre-manipulation trajectories through field-guided planning, enabling scalable diversity generation.

\subsection{Automated and Simulation-Based Data Generation}

To generate large-scale data with minimal human input, leveraging simulation engines to assist in data synthesis has become an emerging trend~\cite{xue2025demogen, hoque2024intervengen, garrett2024skillmimicgen, yu2025skillmimic, wang2023gensim}. Large-scale simulation platforms like RoboTwin~\cite{mu2024robotwin}, MimicGen~\cite{mandlekar2023mimicgen}, and DexMimicGen~\cite{jiang2025dexmimicgen} can generate millions of diverse manipulation scenarios through procedural generation and domain randomization. These systems achieve unprecedented scale by parallelizing data collection across multiple simulated environments and automatically varying object properties, lighting conditions, and scene configurations.
Recent advances have pushed simulation-based generation to new scales. RoboTwin 2.0~\cite{chen2025robotwin} provides a comprehensive benchmark with diverse manipulation tasks, while HumanoidGen~\cite{jing2025humanoidgendatagenerationbimanual} focuses specifically on bimanual dexterous manipulation. These platforms demonstrate the potential for generating vast quantities of training data that would be impossible to collect through teleoperation alone.
However, simulation-based approaches face several fundamental limitations that constrain their effectiveness. First, the persistent sim-to-real gap remains a critical challenge, as policies trained exclusively on synthetic data often fail to transfer robustly to physical systems due to discrepancies in contact dynamics, material properties, and sensor characteristics. Second, purely simulation-generated data suffers from limited behavioral diversity, as it primarily relies on procedurally-generated trajectories that lack the rich variability of contact interactions and manipulation strategies observed in human behaviors. These constraints fundamentally limit the ability of simulation-only approaches to capture the full complexity of dexterous manipulation tasks. 

Our approach sidesteps the sim-to-real problem by generating data directly in the real world. By automating the pre-manipulation phase while relying on human demonstration for contact-rich interactions, FieldGen combines automated scalability with real-world fidelity.

\section{METHOD}

\subsection{Problem Formulation}
We formulate the problem as collecting or generating pairs of observations $o_t$ and action sequences $a_t$. To this end, we divide the data collection process into two phases: (1) Fine manipulation phase. In this stage, we use teleoperation to annotate the manipulation pose. This pose are further used to construct the pre-manipulation field $\mathcal{F}_{Gen}$, an abstract representation of the demonstrated trajectories. (2) FieldGen-based reach phase. In this stage, we employ FieldGen to conduct large-scale, semi-automated data collection. Randomized observations are sampled from diverse end-effector configurations, and corresponding action sequences are generated by asymptotically rolling out toward the pre-manipulation field.

This formulation ensures that fine manipulation is supervised by human expertise, while large-scale reach trajectories are automatically synthesized, providing broad coverage and scalable data generation.

\subsection{$\mathcal{F}_{Gen}$ Field Construction}

\begin{figure}[!t] 
  \centering

  \begin{subfigure}{0.25\textwidth}
    \centering
    \includegraphics[width=\linewidth]{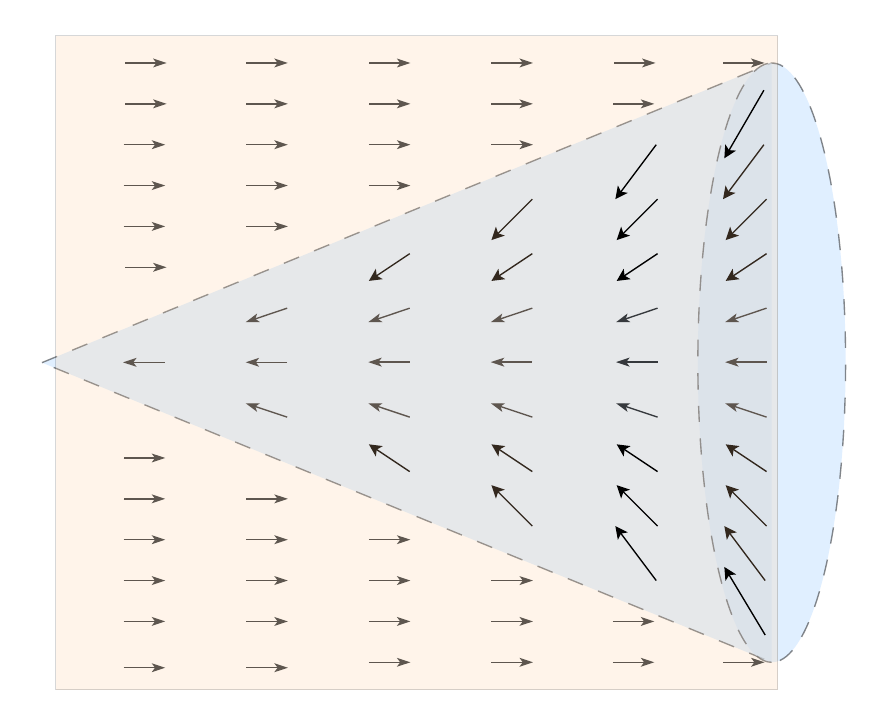}
    \caption{}
    \label{fig2:cone_field}
  \end{subfigure}
  \hfill
  \begin{subfigure}{0.21\textwidth}
    \centering
    \includegraphics[width=\linewidth]{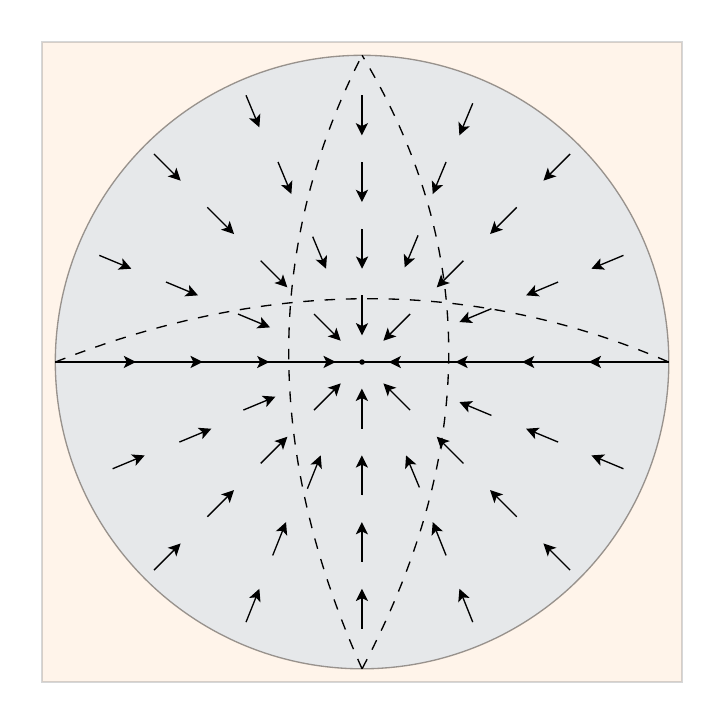}
    \caption{}
    \label{fig2:spherical_field}
  \end{subfigure}

  \caption{(a) Cone Field for Position. (b) Spherical Field for Orientation.}
\end{figure}

In this part, we use the previously collected fine manipulation trajectories to construct the pre-manipulation field $\mathcal{F}_{Gen}$. Based on this field, automated scripts are employed to randomly drive the robot arm, during which large amounts of end-effector poses and corresponding RGB observations are recorded. For each sampled pose, we compute a reach action sequence by evaluating its spatial relationship to the field and solving the corresponding inverse kinematics (IK). In this way, we generate observation–action pairs that serve as training data.

We decompose the generated field $\mathcal{F}_{Gen}$ into two components: 
a cone field for position, $\mathcal{F}_{pos}$, and a spherical field for orientation, $\mathcal{F}_{ori}$. 
The position field captures how the end-effector should approach the manipulation point, 
while the orientation field ensures alignment of the gripper with the desired manipulation orientation.

\subsubsection{Cone Field for Position}

Intuitively, the cone field constrains the end-effector to approach the manipulation position in alignment with the gripper axis. 
When the end-effector lies within the cone, the motion smoothly converges toward the target. 
When outside, the motion is first redirected into the cone and then guided to the manipulation point, 
mimicking a natural reach-and-align strategy.

Formally, let the manipulation goal position be $p_G \in \mathbb{R}^3$, 
with axis direction $\hat{u}$ (opposite to the gripper closing axis) and cone half-angle $\theta$. 
For a sampled point $Q$ with position $p_Q$, we decompose
\begin{equation}
a = \hat{u}^\top (p_Q - p_G), 
\qquad 
r = \big\|(p_Q - p_G) - a\hat{u} \big\|.
\end{equation}
The cone surface is given by
\begin{equation}
r = \tan\theta \, a, \quad a \geq 0.
\end{equation}

As illustrated in Figure~\ref{fig2:cone_field}, the cone field is structured such that $G$ acts as a zero-gravity sink, with field lines converging smoothly toward $G$ within the cone and aligning parallel to $\hat{u}$ outside the cone.

Under the influence of the cone field, if $Q$ lies inside the cone ($r \leq \tan\theta \, a$), the attraction vector is defined by a smooth half-cycloid curve in the plane spanned by $(G, Q, \hat{u})$:
\begin{equation}
x(t) = \mu (t - \sin t), 
\qquad
y(t) = \nu (1 - \cos t), 
\qquad
t \in [0, \pi],
\end{equation}
with $\mu, \nu$ chosen so that the curve starts at $Q$ and ends at $G$.  
If $Q$ lies outside the cone ($r > \tan\theta \, a$), it is first projected along $\hat{u}$ onto the cone surface at $P$, then follows the inner cycloid curve $P \to G$.  

Thus the translational delta action is
\begin{equation}
\Delta p = \operatorname{Curve}(p_Q \to p_G \mid \hat{u}, \theta).
\end{equation}

\subsubsection{Spherical Field for Orientation}

The spherical field ensures direct alignment of the gripper with the target manipulation orientation. The three axes of the spherical field correspond to roll, pitch, and yaw, with all field lines converging toward the center of the sphere as shown in Figure~\ref{fig2:spherical_field}.
It provides a smooth corrective rotation that gradually drives the end-effector toward the goal orientation $R_G$.

Let $R_G \in SO(3)$ denote the goal orientation and $R_Q$ the sampled orientation.  
The relative rotation is
\begin{equation}
R_\Delta = R_G^\top R_Q.
\end{equation}
The axis--angle representation is
\begin{equation}
\omega = \log(R_\Delta) \in \mathbb{R}^3,
\end{equation}
and the corrective angular velocity is defined as
\begin{equation}
\Delta R = -K_R \, \omega.
\end{equation}

After obtaining the generated trajectory based on these fields, the length of action sequence is determined by dividing the trajectory length by the parameter $\beta$. Finally, we extract the action sequence of one chunk-sized segment; if the sequence length is shorter than the chunk size, we pad it with the last point of the trajectory.

\subsection{FieldGen-Reward: Generate Diverse Reward-Annotated Trajectories}

The core principle of \textbf{\textit{FieldGen-Reward}} is to generate a rich dataset where each trajectory is associated with a continuous reward value. Specifically, for a given successful manipulation endpoint $P_O$ from teleoperation, we define a sphere of radius $R$ centered at this point. We then randomly sample a new endpoint $P_N$ within this sphere. A new trajectory is generated towards this new endpoint, and its quality is quantified by a reward signal calculated as below, where d is the Euclidean distance between the original and the new endpoint:

\begin{equation}
d=|P_OP_N|,
\qquad
reward=1-d/R
\end{equation}

This formulation creates a continuous distribution of data quality, from perfect trajectories (reward of 1, when $d=0$) to less precise ones (reward approaching 0, as $d$ approaches $R$). This method theoretically allows for the generation of infinite trajectories, each with a corresponding reward.
\section{EXPERIMENT}

\subsection{Objective and Setup} 
We design experiments to evaluate FieldGen along three questions:

(1) Under the same collection time budgets, whether FieldGen produces training data that yields stronger policy performance than purely teleoperated data?

(2) At a fixed dataset size, whether FieldGen provides higher per-sample data quality, assessed by success rate and generalization across tasks?

(3) Through ablation studies, whether each component of FieldGen contributes essentially to overall effectiveness.

Experiments were conducted on the Agibot G1 robot, using an NVIDIA Orin for inference. To comprehensively evaluate the effect of data distribution and generation methods, we train all policies from scratch, removing any influence from pretrained data. We compare three policies: small RDT~\cite{liu2024rdt} (170M, SigLIP~\cite{zhai2023sigmoid} frozen encoder), DP~\cite{chi2023diffusion} (288M, jointly trained encoder and head) and ACT~\cite{zhao2023learning}. Since ACT achieved consistently low success rates across most tasks, its results are omitted from the tables for clarity. Observations are wrist-mounted RGB images; actions are end-effector delta poses. Each action chunk outputs 30 steps.

\subsection{Equal-Time Data Effectiveness}

\begin{figure}[h]
    \centering
    \captionsetup[sub]{font=small,justification=centering}

    \begin{subfigure}[t]{0.48\linewidth}
        \centering
        \includegraphics[width=\linewidth, keepaspectratio]{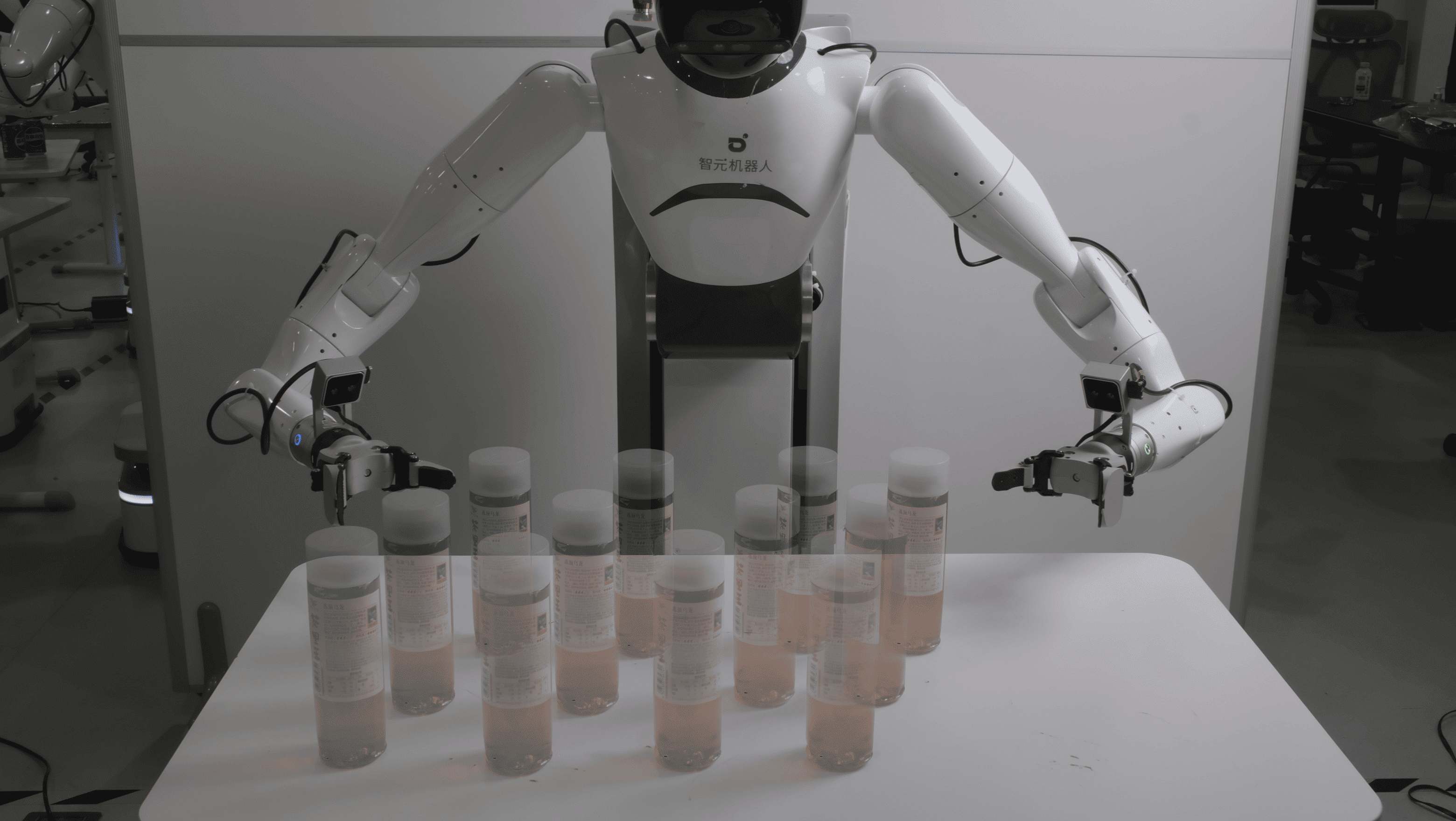}
        \caption{Pick}
        \label{fig:setup_pick}
    \end{subfigure}\hfill
    \begin{subfigure}[t]{0.48\linewidth}
        \centering
        \includegraphics[width=\linewidth, keepaspectratio]{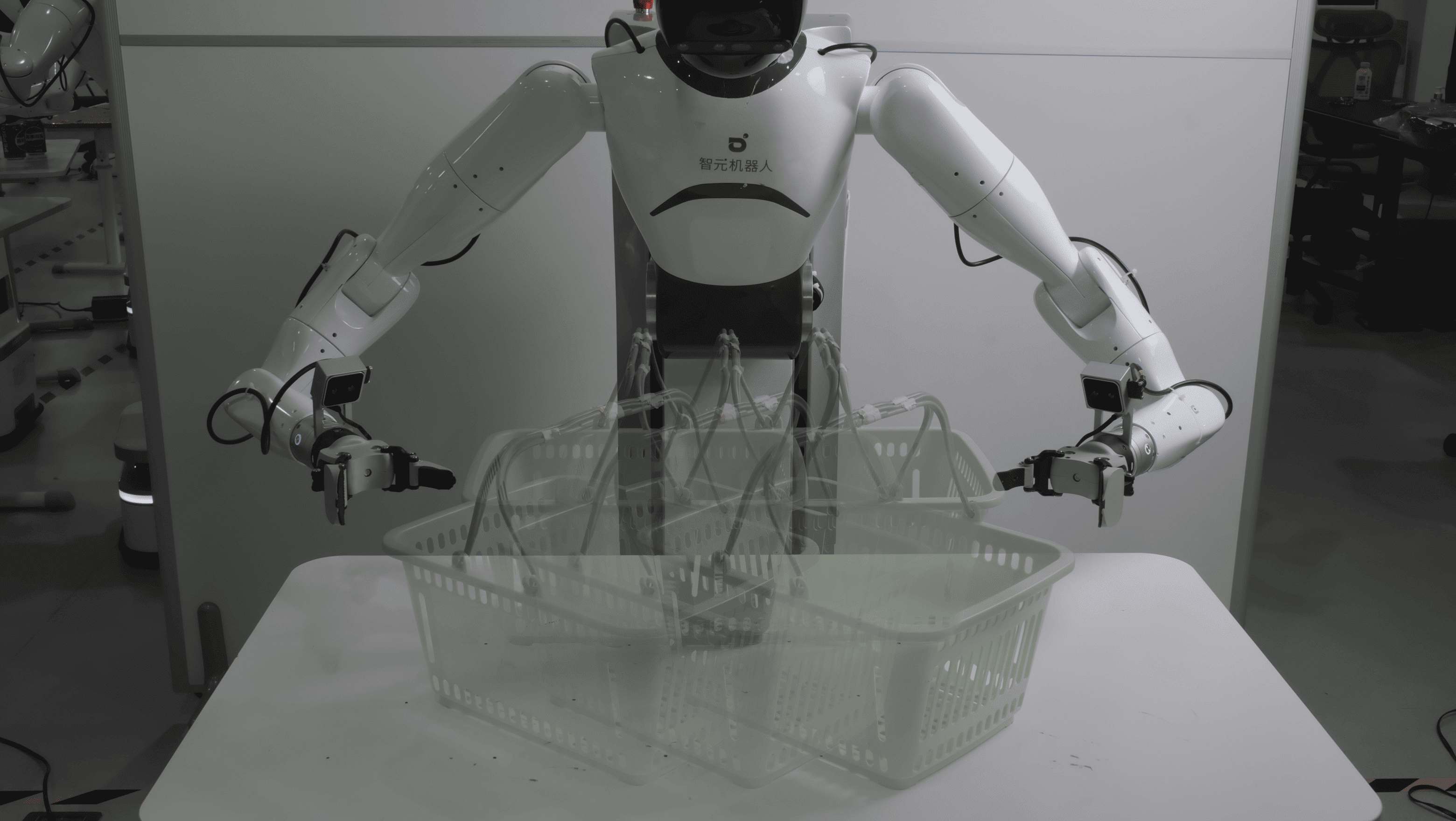}
        \caption{Rotate Pick}
        \label{fig:setup_rotatepick}
    \end{subfigure}

    \vspace{4pt}

    \begin{subfigure}[t]{0.48\linewidth}
        \centering
        \includegraphics[width=\linewidth, keepaspectratio]{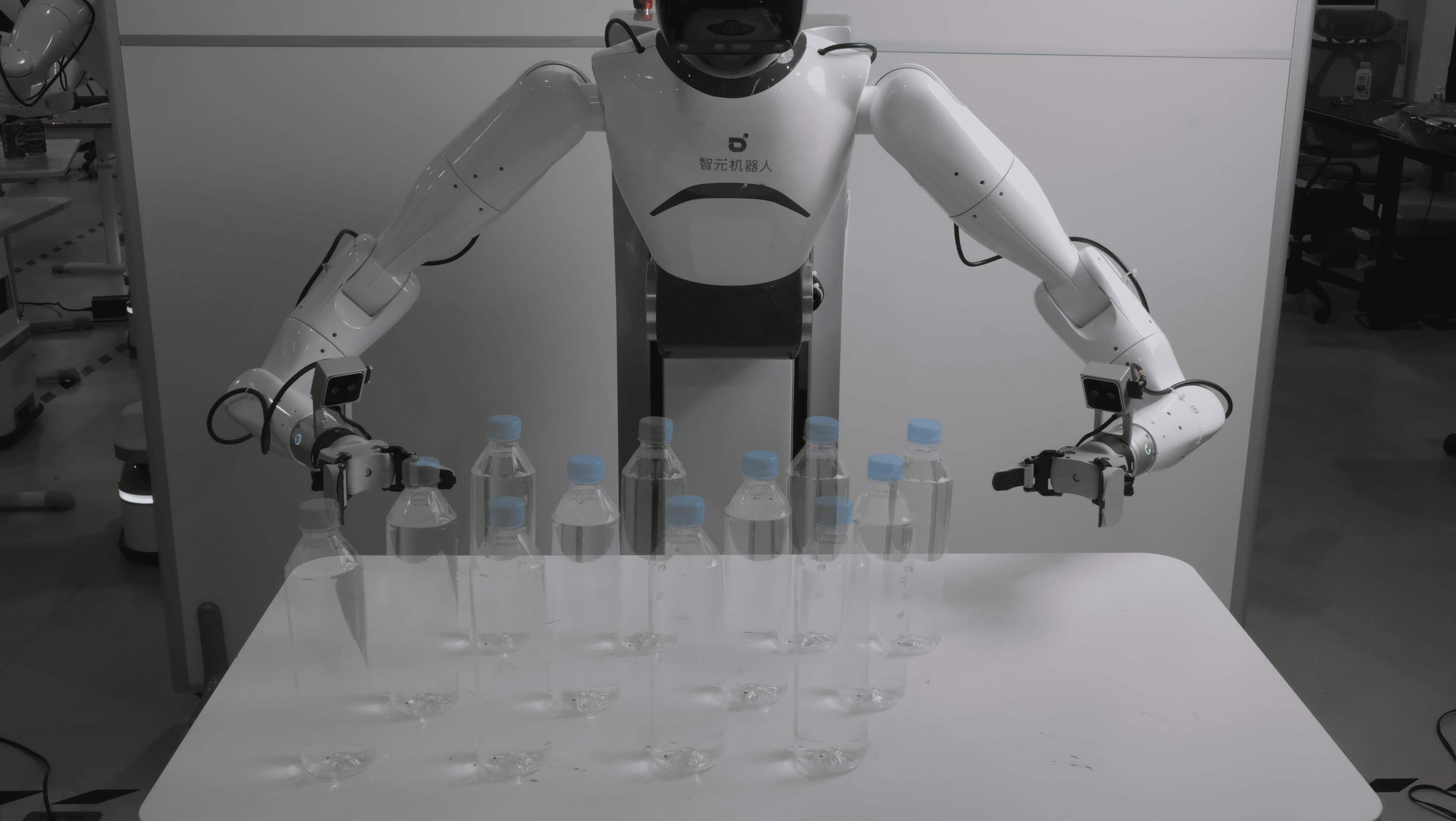}
        \caption{Transparent Pick}
        \label{fig:setup_transparentpick}
    \end{subfigure}\hfill
    \begin{subfigure}[t]{0.48\linewidth}
        \centering
        \includegraphics[width=\linewidth, keepaspectratio]{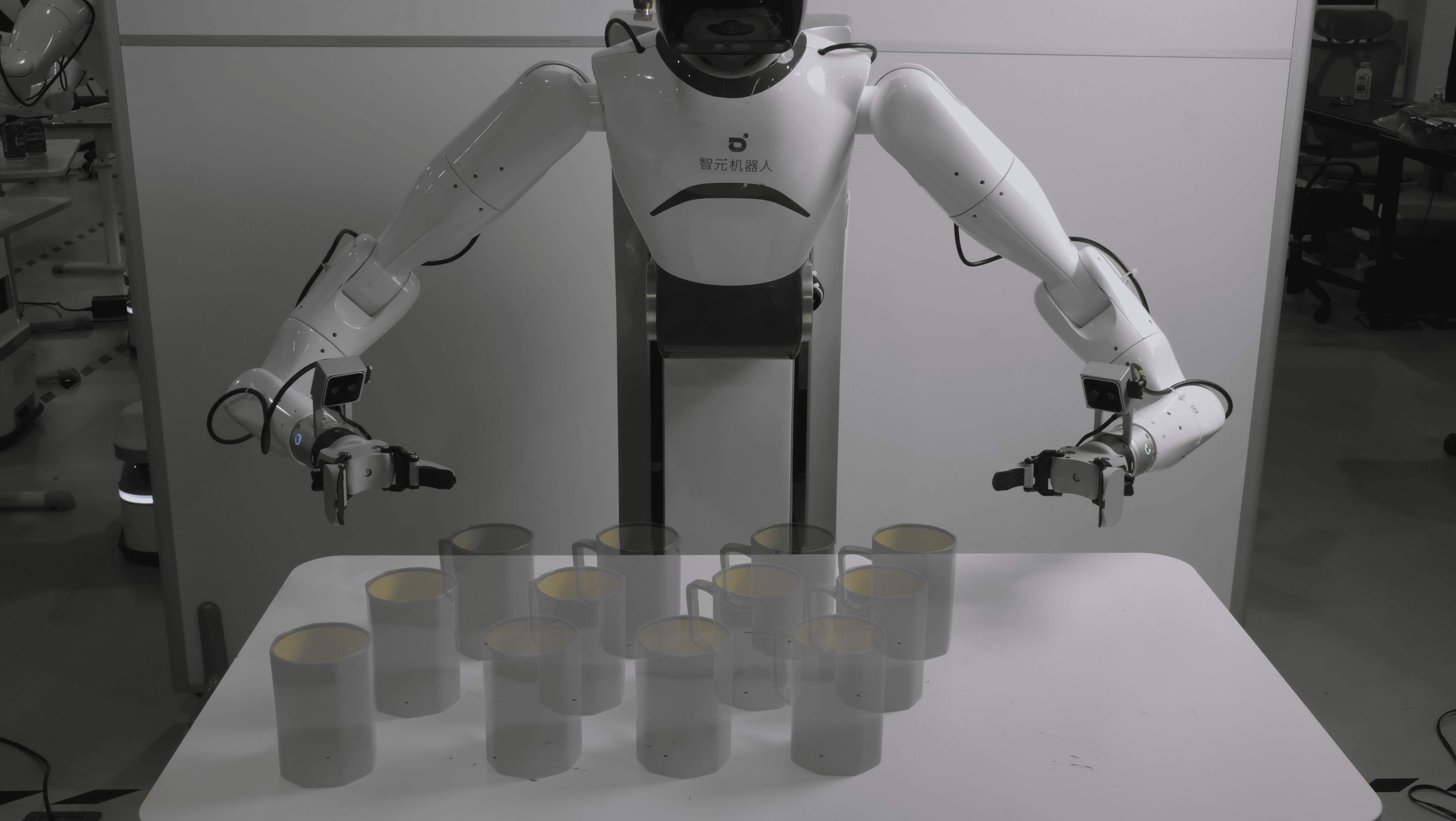}
        \caption{Affordance Pick}
        \label{fig:setup_affordancepick}
    \end{subfigure}

    \vspace{-2pt}
    \caption{\textbf{Equal-Time Data Effectiveness Experiment Setups.} Rotate Pick requires the end-effector change orientation to pick objects, while Affordance Pick requires to manipulate specific position.}
    \label{fig:exp1_setups}
\end{figure}

\begin{table*}[!t]
\centering
\footnotesize
\setlength{\tabcolsep}{4pt}
\captionof{table}{\textbf{Equal-Time Data Effectiveness.} Rows list tasks (DP/RDT-small); columns list Teleop vs. FieldGen across collection times.}
\label{tab:equal-time-reorg}
\begin{tabular}{lccccccccccc}
    \toprule
    \multirow{1}{*}{Task} & \multirow{1}{*}{Model} &
    \multicolumn{5}{c}{Teleop} & \multicolumn{5}{c}{FieldGen} \\
    \cmidrule(lr){3-7} \cmidrule(lr){8-12}

    \multirow{1}{*}{Collect Time(min)}
      &  & 4 & 8 & 12 & 16 & 20 & 4 & 8 & 12 & 16 & 20 \\
    
    \midrule
    \multirow{2}{*}{Pick}
      & DP  & 16.7\% & 33.3\% & 58.3\% & 66.7\% & 83.3\% & 75.0\% & 91.7\% & 91.7\% & 83.3\% & 91.7\% \\
      & RDT-small & 16.7\% &  8.3\% & 33.3\% & 16.7\% & 25.0\% & 58.3\% & 83.3\% & 83.3\% & 83.3\% & 91.7\% \\
      \cmidrule(lr){2-12}
    \multirow{2}{*}{Rotate Pick}
      & DP  & 25.0\% & 33.3\% & 66.7\% & 75.0\% & 75.0\% & 58.3\% & 66.7\% & 91.7\% & 91.7\% & 100\% \\
      & RDT-small &  8.3\% & 41.7\% & 50.0\% & 58.3\% & 83.3\% & 41.7\% & 66.7\% & 91.7\% & 100\%  & 100\% \\
      \cmidrule(lr){2-12}
    \multirow{2}{*}{Transparent Pick}
      & DP  &  8.3\% & 16.7\% &  8.3\% & 16.7\% &  8.3\% & 58.3\% & 91.7\% & 100\%  & 100\%  & 100\% \\
      & RDT-small &  8.3\% & 25.0\% & 33.3\% & 25.0\% & 41.6\% & 83.3\% & 75.0\% & 83.3\% & 83.3\% & 83.3\% \\
      \cmidrule(lr){2-12}
    \multirow{2}{*}{Affordance Pick}
      & DP  & 66.7\% & 75.0\% & 75.0\% & 83.3\% & 91.7\% & 83.3\% & 100\%  & 100\%  & 100\%  & 100\% \\
      & RDT-small & 33.3\% & 58.3\% & 50.0\% & 66.7\% & 75.0\% & 58.3\% & 75.0\% & 100\%  & 100\%  & 100\% \\

    \midrule
    \multirow{2}{*}{Average}
      & DP  & 29.2\% & 39.6\% & 52.1\% & 60.4\% & 64.6\% & 68.8\% & 87.5\% & 95.8\% & 93.8\% & 97.9\% \\
      & RDT-small & 16.7\% & 33.3\% & 41.7\% & 41.7\% & 56.2\% & 60.4\% & 75.0\% & 89.6\% & 91.7\% & 93.8\% \\
    \bottomrule
\end{tabular}
\end{table*}

\begin{figure*}[!t] 
  \centering

  \begin{subfigure}{0.53\textwidth}
    \centering
    \includegraphics[width=\linewidth]{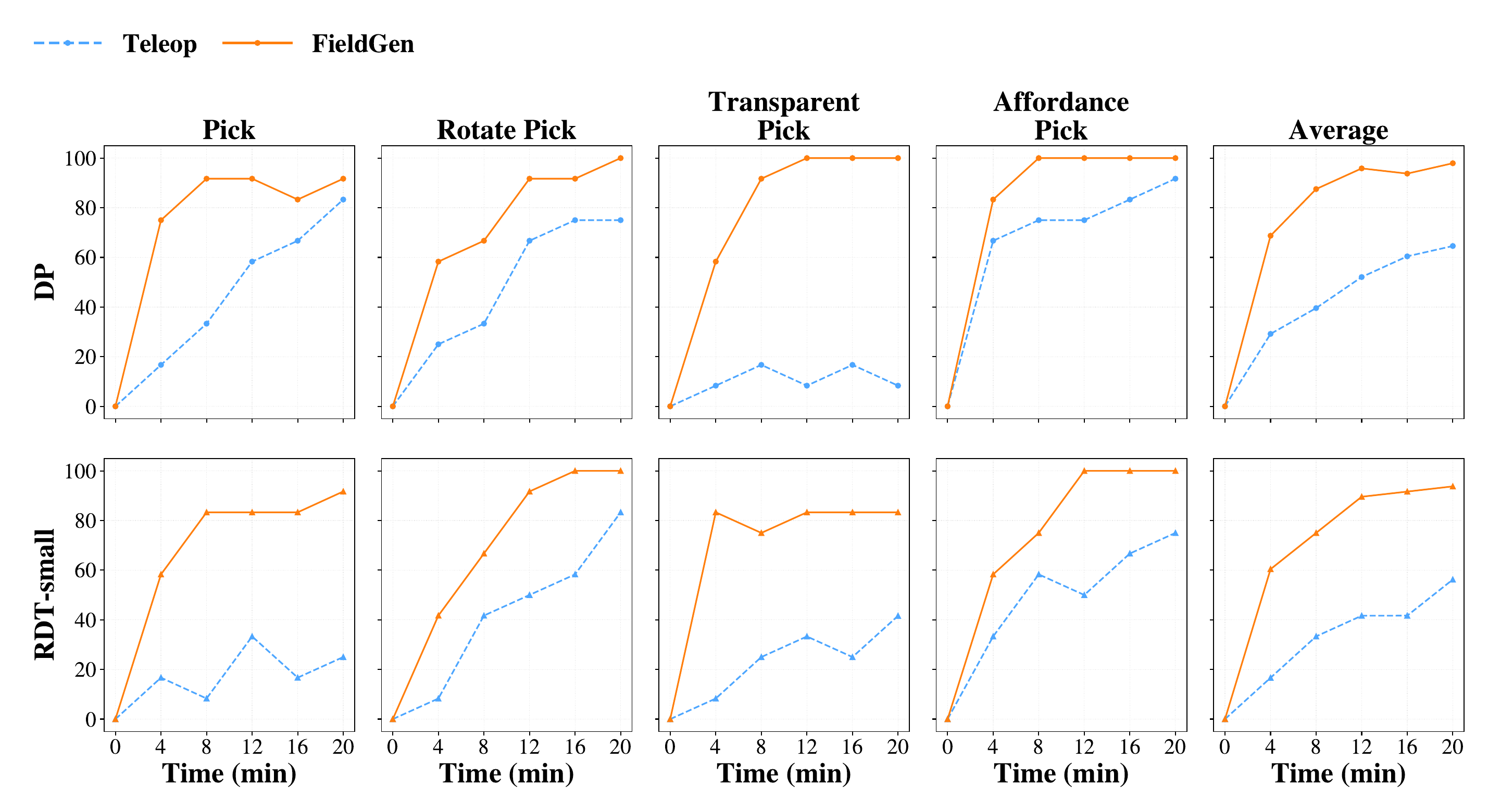}
    \caption{}
    \label{fig1:experiment-1-curve}
  \end{subfigure}
  \hfill
  \begin{subfigure}{0.44\textwidth}
    \centering
    \includegraphics[width=\linewidth]{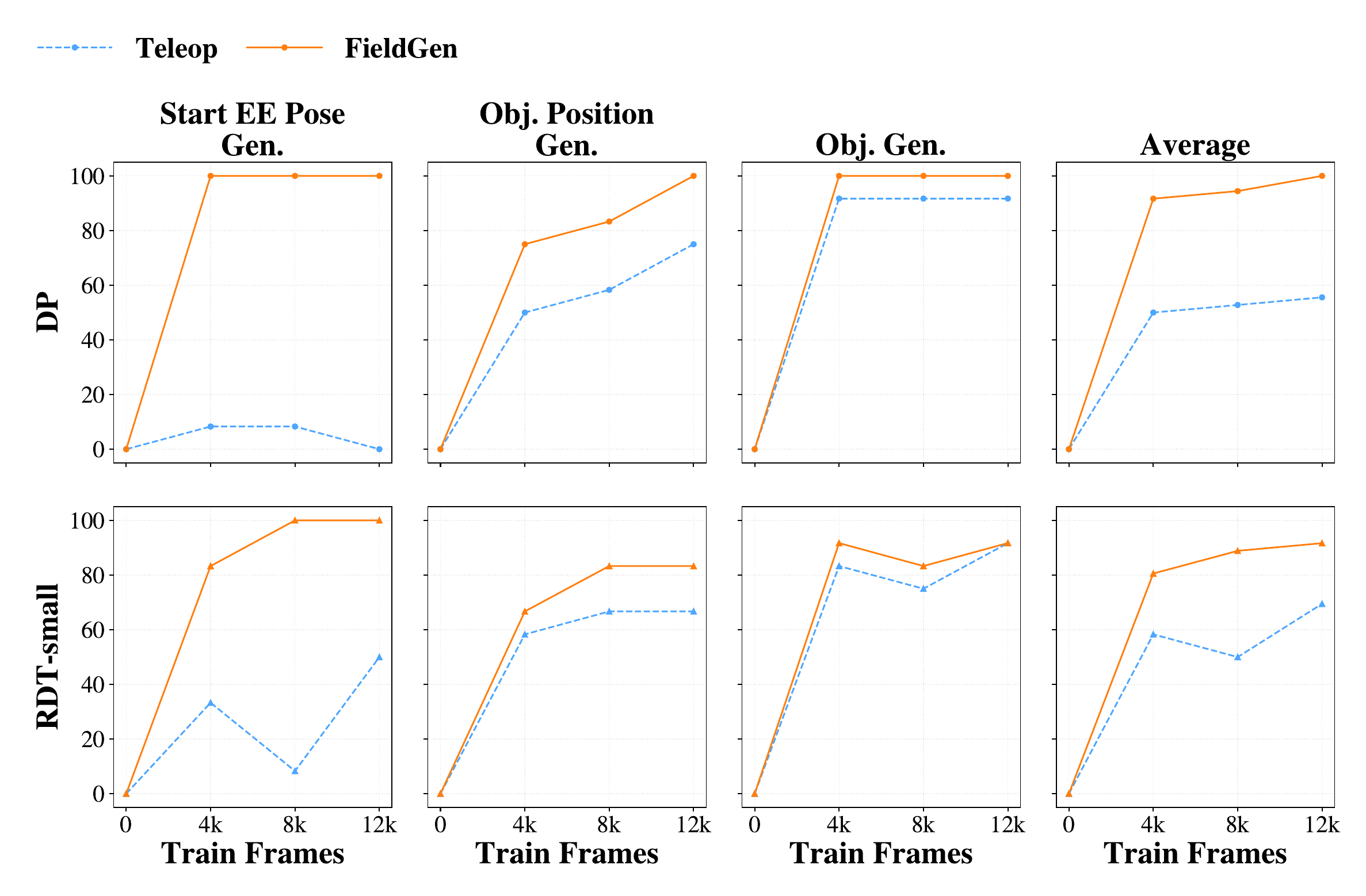}
    \caption{}
    \label{fig1:exp2-curve}
  \end{subfigure}

  \caption{(a) Equal-Time Data Effectiveness. (b) Equal-Data Generalization.}
\end{figure*}

To verify the efficiency of FieldGen, we design experiments that compare policies trained on data collected by FieldGen with those trained on fully teleoperated data, under equal wall-clock collection budgets. Specifically, data are collected for checkpoints every 4 minutes up to 20 minutes. This setup also allows us to examine whether increasing data scale correlates with improved policy success rates. For each task, FieldGen collects a frame of the manipulation pose and then automatically gathers randomized observations until each episode reaches 1 minute in duration. As shown in Figure~\ref{fig:exp1_setups}, we evaluate four manipulation tasks: Pick, Rotate Pick, Transparent Pick, and Affordance Pick. Data are obtained either through full teleoperation or through FieldGen. All policies are trained for 50 epochs until convergence, and success rates are measured across 12 randomized object placements per task.

From the results in Table~\ref{tab:equal-time-reorg}, we observe that FieldGen consistently outperforms teleoperation across all policies and all time budgets. At each checkpoint (4–20 minutes), FieldGen achieves higher success rates, exceeding teleoperation by 41.7\%, 44.8\%, 45.8\%, 41.7\%, and 35.5\% respectively. This highlights the strong time efficiency of FieldGen: under equal wall-clock collection budgets, the data it generates leads to substantially stronger policy performance.

Figure~\ref{fig1:experiment-1-curve} further shows that FieldGen scales more effectively with additional collection time. Success rates grow more sharply with time compared to teleoperation, and after 20 minutes of collection, FieldGen exceeds 80\% success in all settings. In particular, diffusion-policy–based methods reach 100\% success on three of the four evaluated tasks, reflecting the high quality of FieldGen data.

\subsection{Data Quality and Generalization}

\begin{figure}[t]
    \centering
    \includegraphics[width=\linewidth]{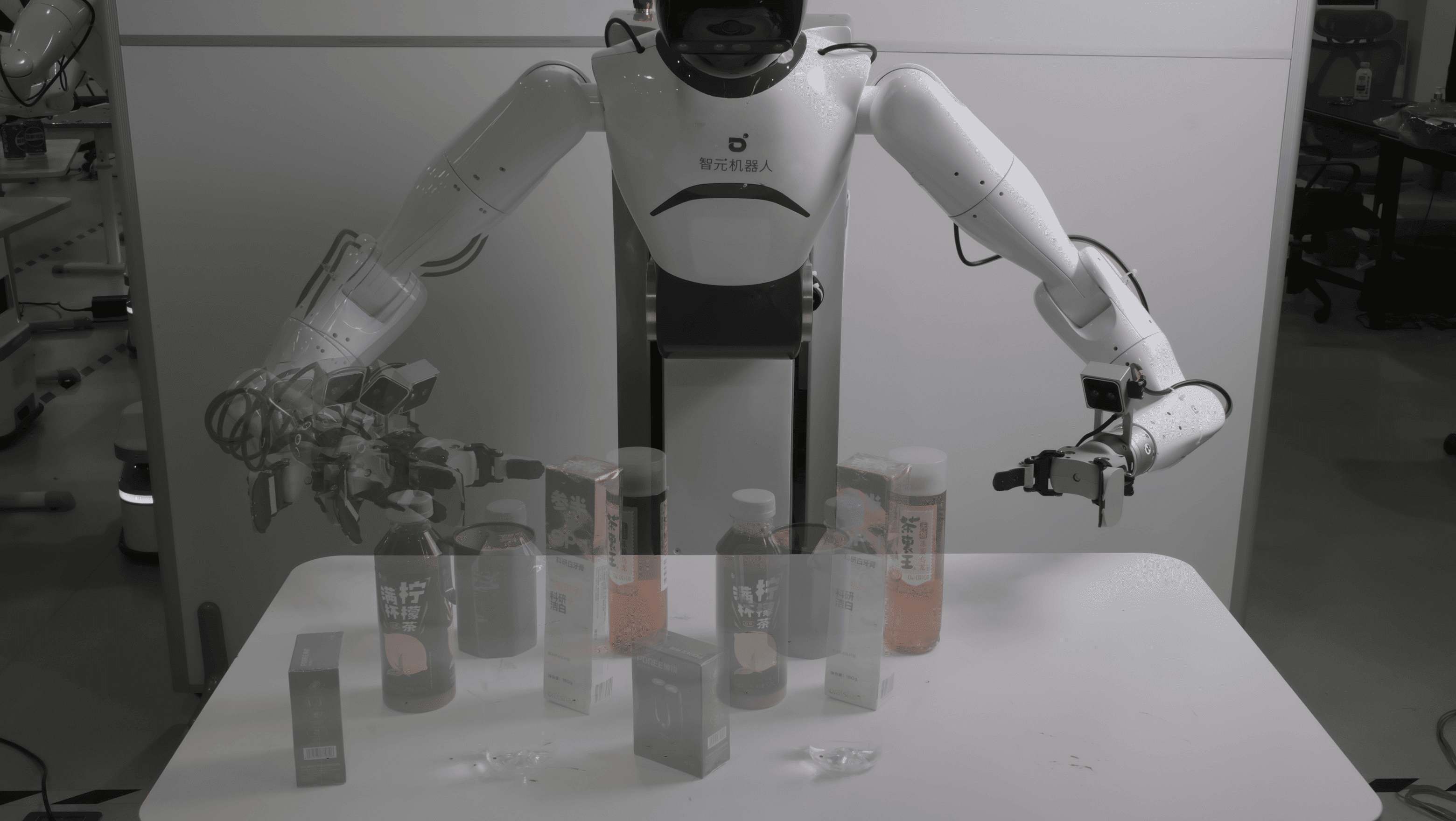}
    \caption{\textbf{Data Quality and Generalization Experiment Setups}}
    \label{fig:exp2_setups}
\end{figure}

\begin{table}[!t]
  \centering
  \footnotesize
  \setlength{\tabcolsep}{2pt}
  \caption{\textbf{Generalization Experiment Results.} We conduct controlled experiments on 3 generalization tasks: \textit{Start EE Pose Gen}, \textit{Obj. Position Gen}, and \textit{Obj. Gen}, each evaluated under 3 different data volume.}
  \begin{tabular}{lccccc}
    \toprule
    \multirow{2}{*}{\textbf{Task}}
    & \multirow{2}{*}{\textbf{Train Frames}}
    & \multicolumn{2}{c}{\textbf{Teleop}}
    & \multicolumn{2}{c}{\textbf{FieldGen}} \\
    \cmidrule(lr){3-4} \cmidrule(lr){5-6}
    & & \textit{DP} & \textit{RDT-small} & \textit{DP} & \textit{RDT-small} \\
    \midrule
    \multirow{3}{*}{Start EE Pose Gen.}
      & 4000  & 8.3\%  & 33.3\% & \textbf{100\%} & \textbf{83.3\%} \\
      & 8000  & 8.3\%  & 8.3\%  & \textbf{100\%}  & \textbf{100\%}  \\
      & 12000 & 0\%    & 50.0\% & \textbf{100\%}  & \textbf{100\%}  \\
      \cmidrule(lr){2-6}
    \multirow{3}{*}{Obj. Position Gen.}
      & 4000  & 50.0\% & 58.3\% & \textbf{75.0\%} & \textbf{66.7\%} \\
      & 8000  & 58.3\% & 66.7\% & \textbf{83.3\%} & \textbf{83.3\%} \\
      & 12000 & 75.0\% & 66.7\% & \textbf{100\%}  & \textbf{83.3\%} \\
      \cmidrule(lr){2-6}
    \multirow{3}{*}{Obj. Gen.}
      & 4000  & 91.7\% & 83.3\% & \textbf{100\%}  & \textbf{91.7\%} \\
      & 8000  & 91.7\% & 75.0\% & \textbf{100\%}  & \textbf{83.3\%} \\
      & 12000 & 91.7\% & 91.7\% & \textbf{100\%}  & \textbf{91.7\%} \\
    \midrule
    \multirow{3}{*}{Average}
      & 4000  & 50.0\% & 58.3\% & \textbf{91.7\%} & \textbf{80.6\%} \\
      & 8000  & 52.8\% & 50.0\% & \textbf{94.4\%} & \textbf{88.9\%} \\
      & 12000 & 55.6\% & 69.5\% & \textbf{100\%}  & \textbf{91.7\%} \\
    \bottomrule
  \end{tabular}
  \label{tab:exp2-generalization}
\end{table}

To verify the quality of FieldGen data, we design experiments that compare its effect on task success rate and generalization. Specifically, we aim to assess whether policies trained on FieldGen data achieve higher success rates than those trained on teleoperation-only datasets of the same size, and whether FieldGen provides stronger generalization across different tasks and scene variations.

In this setup, we consider three evaluation conditions: (1) end-effector initialization, where policies are tested under diverse initial poses of the robot end-effector; (2) object placement, where policies are evaluated on varied object positions within the workspace; and (3) cross-instance generalization, where policies are transferred to unseen object instances within the same category, as shown in Figure~\ref{fig:exp2_setups}. For both teleoperation and FieldGen, datasets of equal size (measured in data frames) are collected and used to train policies. All models are trained under identical schedules and hyperparameters. Success rates are measured across multiple randomized trials for each condition, enabling a direct comparison of robustness and generalization between the two data sources.

The results are summarized in Table~\ref{tab:exp2-generalization}. Under equal dataset sizes (data frames), FieldGen consistently outperforms teleoperation across both policy backbones. Notably, for diffusion policy (DP), only 4000 FieldGen samples are sufficient to reach 100\% success on both the Start EE Pose Generalization and Object Generalization tasks.

We further plot success rate against dataset size in Figure~\ref{fig1:exp2-curve}. FieldGen shows a stable upward trend that follows a clear scaling curve, while teleoperation data consistently lags behind and exhibits irregular growth. This indicates that FieldGen covers a broader range of task-relevant state space, enabling more stable training and stronger performance scaling. In contrast, teleoperation trajectories are subject to higher randomness in human motion, resulting in narrower coverage and more variable downstream performance.

\subsection{Trajectory Diversity and Spatial Coverage}

To evaluate the trajectory diversity of FieldGen, we designed experiments that compare its spatial coverage against teleoperated data. Specifically, we considered three levels of data diversity corresponding to different collection strategies. The first setting uses teleoperation with the same initial and final end-effector states, which yields relatively uniform trajectories and is referred to as low diversity. The second setting still relies on teleoperation but allows varied initial end-effector states while keeping the final object configuration fixed, thereby producing moderately diverse trajectories, denoted as middle diversity. The third setting applies the FieldGen method, which automatically generates trajectories that span a wide range of spatial variations, achieving high diversity.

\begin{table}[ht]
  \centering
  \footnotesize
  \caption{\textbf{Diversity Comparison.} We compare the spatial coverage and model performance across three different diversity levels.}
  \begin{tabular}{lccc}
    \toprule
    \textbf{Diversity Level} & \textbf{Spatial Coverage} & \textbf{DP} & \textbf{RDT-small} \\
    \midrule
    Low    & 9.04\%   & 0\%    & 0\%   \\
    Middle & 15.44\%  & 66.7\% & 41.7\% \\
    High   & 18.14\%  & 83.3\% & 83.3\% \\
    \bottomrule
  \end{tabular}
  \label{tab:exp3-diversity}
\end{table}

\begin{figure*}[!t]
  \centering
  \begin{subfigure}{0.25\textwidth}
    \centering
    \includegraphics[width=\linewidth]{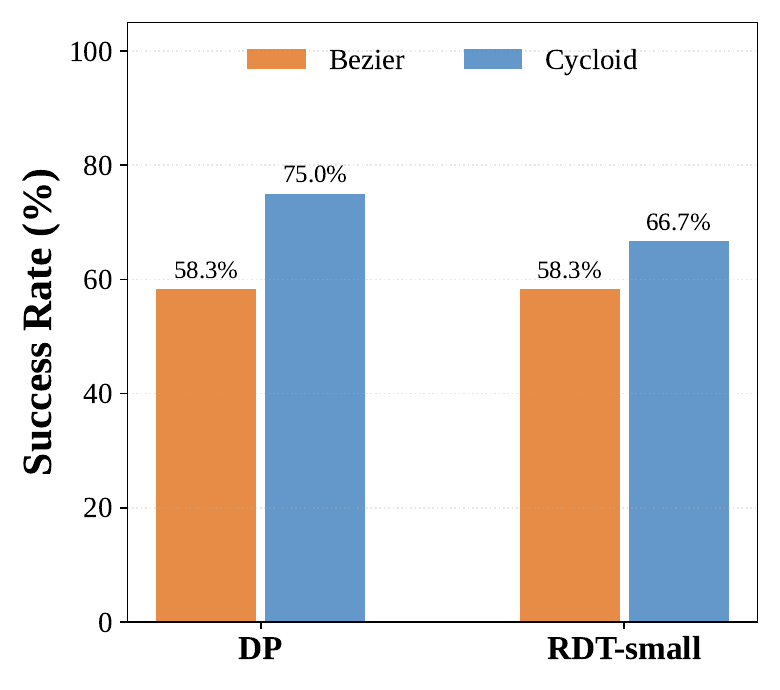}
    \caption{}
    \label{fig:abla-curve}
  \end{subfigure}
  \quad  
  \begin{subfigure}{0.25\textwidth}
    \centering
    \includegraphics[width=\linewidth]{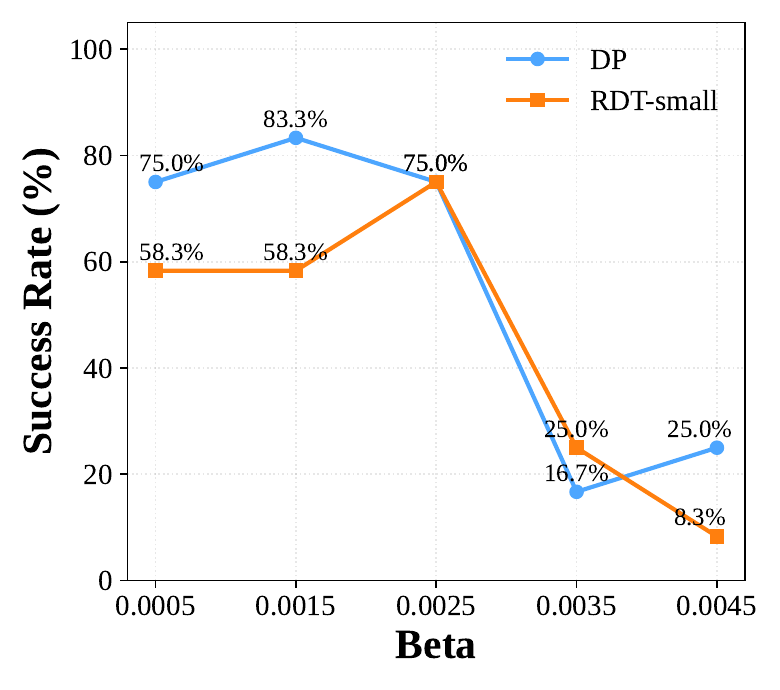}
    \caption{}
    \label{fig:abla-beta}
  \end{subfigure}
  \quad
  \begin{subfigure}{0.25\textwidth}
    \centering
    \includegraphics[width=\linewidth]{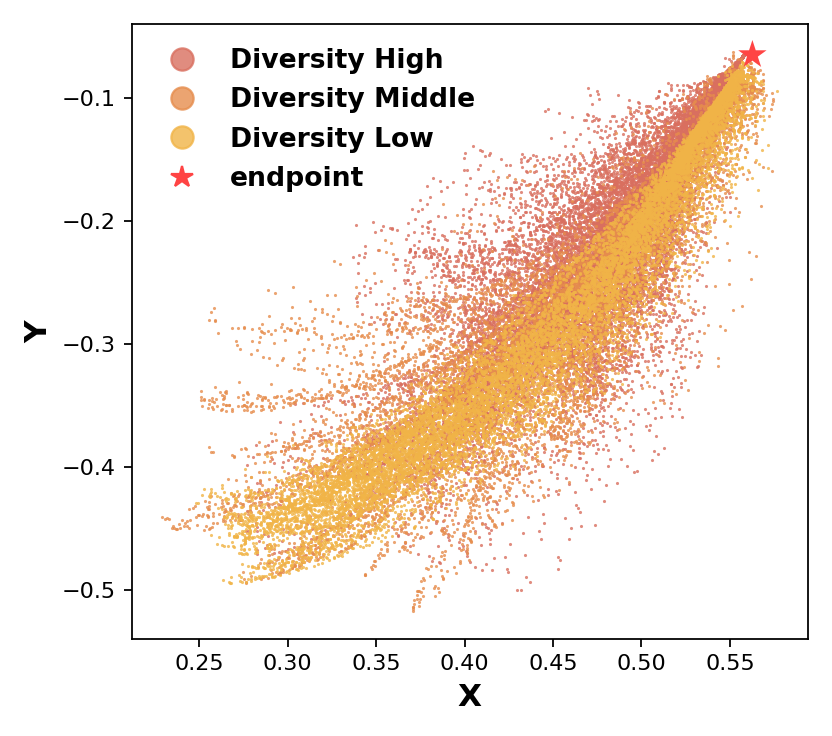}
    \caption{}
    \label{fig1:xyz_diversity}
  \end{subfigure}

  \caption{(a) Ablation on Curve. (b) Ablation on $\beta$ Parameter. (c) $XY$ plane scatter plot of Diversity.}
\end{figure*}

During evaluation, the target object’s spatial position is varied to test policy generalization and robustness. Figure~\ref{fig1:xyz_diversity} illustrates the scatter plot of the collected trajectories along the $XY$ plane, showing that FieldGen produces a broader and more uniform coverage. To quantify this, we calculate spatial coverage by enclosing all trajectories within the minimum bounding cube, dividing the cube into $N$ voxels of equal size, and measuring the fraction $N'/N$ of voxels that are traversed by trajectories. As reported in Table~\ref{tab:exp3-diversity}, FieldGen achieves the highest coverage rate, confirming its ability to explore a broader range of the workspace.

We further compare downstream policy performance using DP and RDT trained on equal-sized datasets collected under the three settings. The resulting success rates are 0\%, 54.2\%, and 83.3\% for low, middle, and high diversity, respectively. These results demonstrate that broader spatial coverage and higher trajectory diversity not only characterize the data itself but also translate directly into more robust and capable manipulation policies. Overall, the experiments verify that FieldGen provides data of higher quality and broader coverage, enabling the training of stronger policies compared to conventional teleoperation.

\subsection{Ablation on Curve Type: B'ezier vs. Cycloid}

We conduct an ablation study to analyze the effect of different curve types on task success rate. In addition to the cycloid curve used in the main approach, we evaluate B'ezier curves—a commonly used alternative in trajectory generation. The results are summarized in Figure~\ref{fig:abla-curve}. Cycloid-based trajectories achieve success rates of 75\% on DP and 66.7\% on RDT, corresponding to improvements of 16.7\% and 8.4\% over B'ezier, respectively. These results highlight the clear advantage of cycloid-based trajectory generation.

To better understand this advantage, we examine the geometric properties of the two approaches. The cycloid is generated directly under the pure geometric constraint of a rolling circle without slipping. This construction provides an explicit geometric prior and yields naturally smooth curvature variations. In contrast, the B'ezier curve relies on manually defined control points to shape the trajectory. Lacking inherent geometric constraints, it often exhibits sharp curvature growth over short distances, resulting in abrupt changes that increase execution difficulty for the manipulator.

Overall, this ablation confirms that cycloid-based generation not only improves task success rates but also aligns with the physical feasibility of robot motion by producing smoother, more geometrically consistent trajectories.

\subsection{Ablation on the $\beta$ Parameter}

We further conduct an ablation study on the parameter $\beta$, which controls the distance between consecutive frames in the generated trajectory. Physically, $\beta$ reflects the average displacement between two successive end-effector poses in Cartesian space. The results on the manipulating task are shown in Figure~\ref{fig:abla-beta}.

When $\beta$ is set to a small value, the inter-frame distance becomes short, causing the manipulator to move only slightly within each chunk. This often leads to repeated local adjustments and oscillatory back-and-forth motions. Moreover, the overall motion speed is reduced, which significantly increases task completion time. In contrast, when $\beta$ is set too large, the inter-frame distance increases, and FieldGen may generate a gripper-closing command while the manipulator is still far from the target object. In real tests, this premature closure prevents accurate manipulating and results in task failure.

Considering both efficiency and reliability, we select $\beta = 0.0025$ as a balanced choice in our experiments.

\subsection{Less Effort with Long-duration Data Collection}

To quantify the differential effort and efficiency between the teleoperation data collection pipeline and the FieldGen generation pipeline, we measured, for a teleoperation operator over a two-hour collection, the proportion of time actually spent performing data collection under different methods, and the resulting throughput, expressed as the rate at which collecting frames outputs were acquired.

We define Collection Time Ratio as the fraction of total collection time during which active data acquisition is occurring. This metric inversely captures procedural and supervisory overhead; higher values therefore indicate improved temporal efficiency. For the FieldGen pipeline, an elevated ratio further signifies diminished operator exertion, since a larger portion of the session consists of autonomous script execution requiring only low‑intensity monitoring. The results are summarized in Figure~\ref{fig7:ratio}. In the experiment, FieldGen yielded a mean Collection Time Ratio of 66.73\%, representing a 2.47× increase over the 27.07\% observed with manual teleoperation, the latter demanding continuous, high‑effort interaction and oversight. Frame collection rate is also as the data generation throughput (frames per minute) of the acquisition process, providing a direct measure of pipeline efficiency. The results are summarized in Figure~\ref{fig7:rate}. FieldGen attained 1203.14 frames/min, exceeding manual teleoperation’s 569.10 frames/min by a factor of 2.11.

\begin{figure*}[t]
    \centering
    \includegraphics[width=\linewidth]{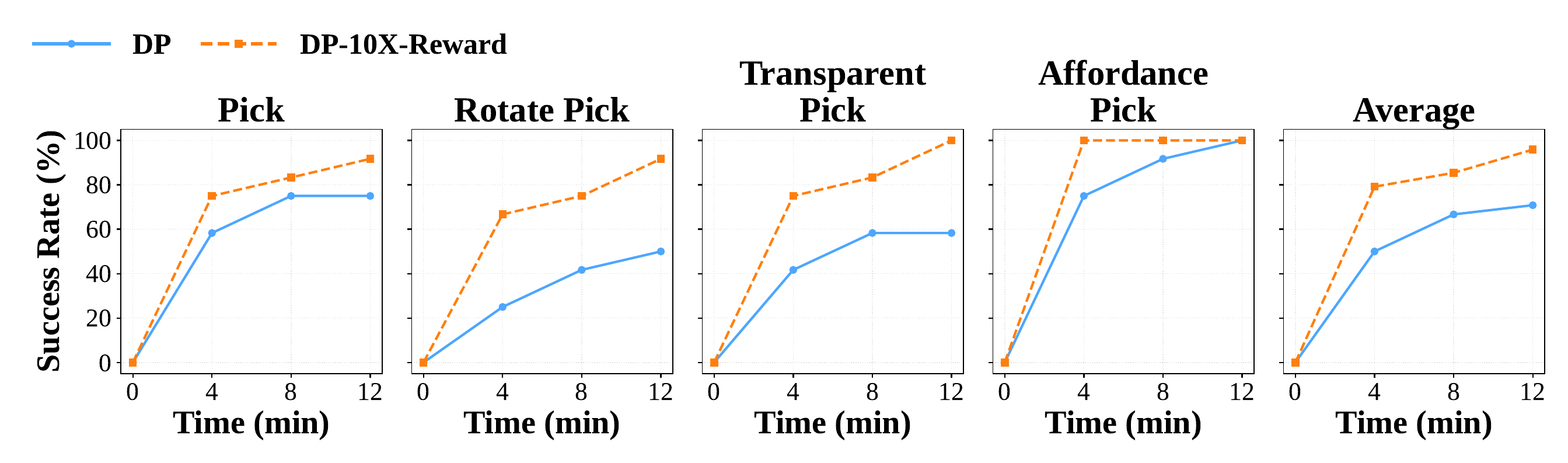}
    \caption{Performance on DP (w/ and w/o FieldGen-Reward).}
    \label{fig:exp_reward}
\end{figure*}

\begin{figure}[!t] 
  \centering

  \begin{subfigure}{0.23\textwidth}
    \centering
    \includegraphics[width=\linewidth]{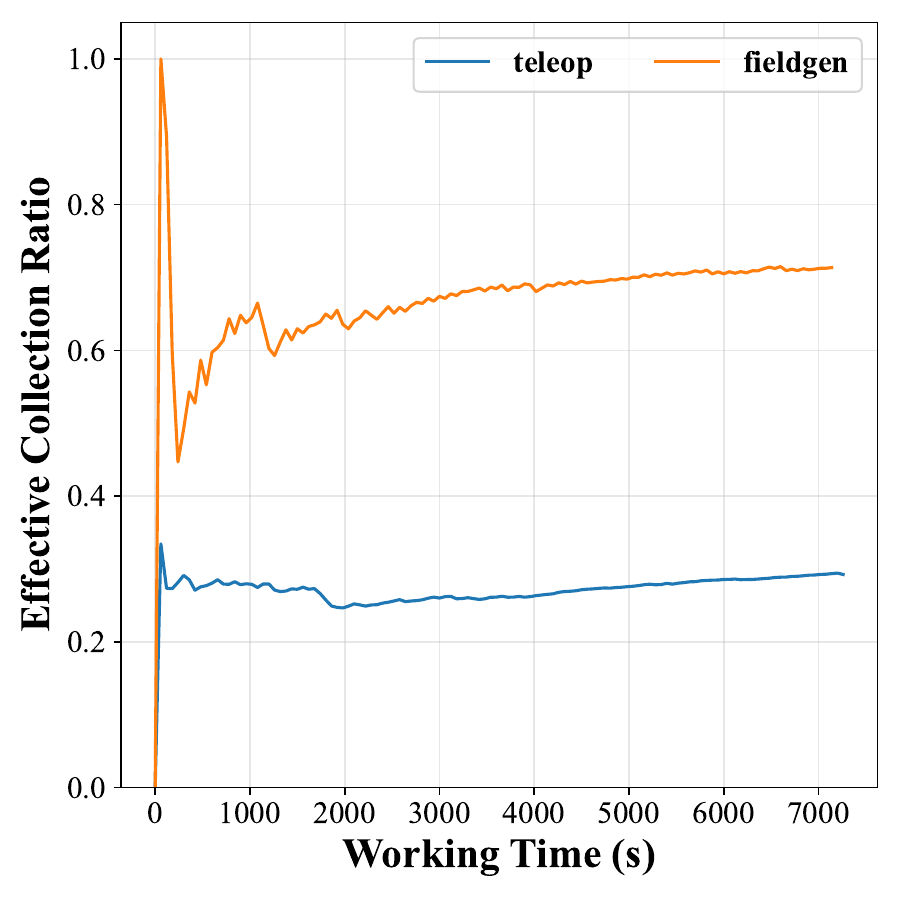}
    \caption{}
    \label{fig7:ratio}
  \end{subfigure}
  \hfill
  \begin{subfigure}{0.23\textwidth}
    \centering
    \includegraphics[width=\linewidth]{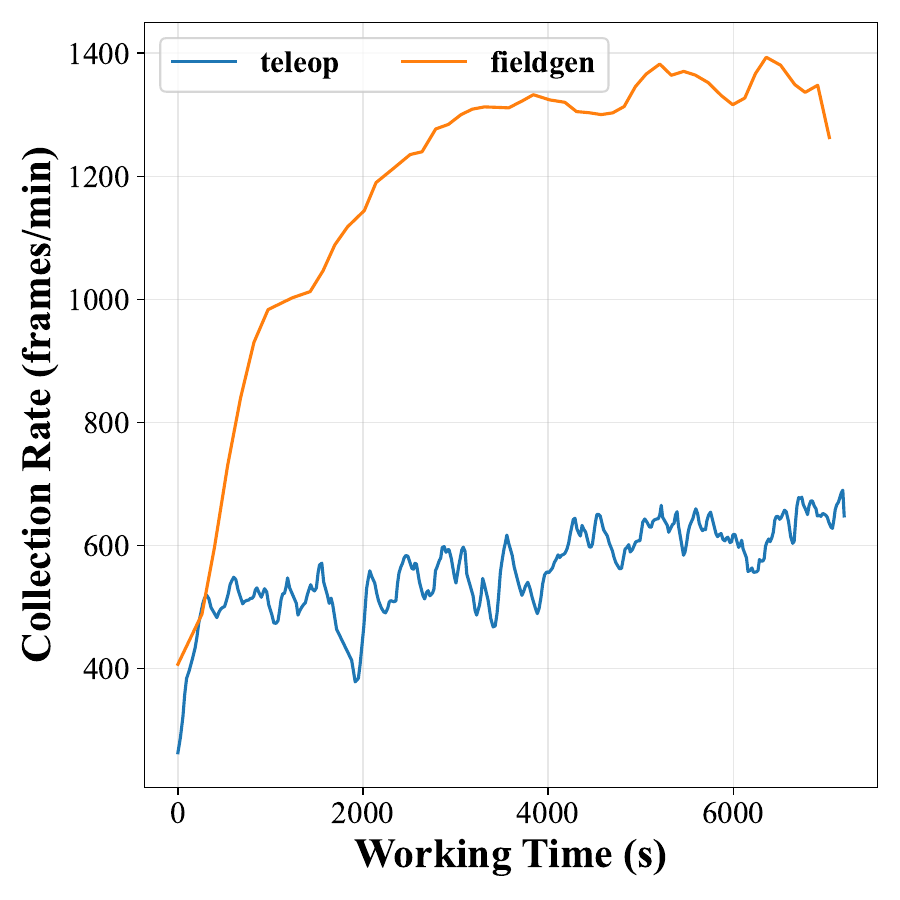}
    \caption{}
    \label{fig7:rate}
  \end{subfigure}

  \caption{(a) Collection time ratio. (b) Memory rate while data collecting.}
\end{figure}

Overall, FieldGen substantially outperforms manual teleoperation in both temporal utilization and data throughput. These experiment results indicate that FieldGen shifts a large fraction of the collection time cost into autonomous execution with only light supervision, thereby reducing operator cognitive and attentional load while simultaneously accelerating scalable dataset production.

\subsection{Ablation on FieldGen-Reward}

\begin{table}[!t]
\centering
\footnotesize
\setlength{\tabcolsep}{4pt}
\captionof{table}{\textbf{Reward Data Effectiveness.} Rows list tasks (DP/RDT-small); columns list FieldGen vs. FieldGen-Reward across coll ection times.}
\label{tab:equal-time-reorg}
\begin{tabular}{lcccc}
    \toprule
    \multirow{1}{*}{Task} & \multirow{1}{*}{Model} &
    \multicolumn{3}{c}{FieldGen} \\
    \cmidrule(lr){3-5} 

    \multirow{1}{*}{Collect Time(min)}
      &  & 4 & 8 & 12   \\
    
    \midrule
    \multirow{2}{*}{Pick}
      & DP  & 58.3\% & 75.0\% & 75.0\% \\
      & DP-R. & 75.0\% & 83.3\% & 91.7\% \\
      \cmidrule(lr){2-5}
    \multirow{2}{*}{Rotate Pick}
      & DP  & 25.0\% & 41.7\% & 50.0\% \\
      & DP-R. & 66.7\% & 75.0\% & 91.7\% \\
      \cmidrule(lr){2-5}
    \multirow{2}{*}{Transparent Pick}
      & DP  & 41.7\% & 58.3\% & 58.3\% \\
      & DP-R. & 75.0\% & 83.3\% & 100.0\% \\
      \cmidrule(lr){2-5}
    \multirow{2}{*}{Affordance Pick}
      & DP  & 75.0\% & 91.7\% & 100.0\% \\
      & DP-R. & 100.0\% & 100.0\% & 100.0\% \\

    \midrule
    \multirow{2}{*}{Average}
      & DP  & 50.0\% & 66.7\% & 70.8\% \\
      & DP-R. & 79.2\% & 85.4\% & 95.9\% \\
    \bottomrule
\end{tabular}
\end{table}

To further enhance data diversity and guide policy learning with explicit quality signals, we introduce a reward-conditioned variant of our method, termed FieldGen-Reward. This approach allows the policy to learn from a broader spectrum of trajectories, including those that are sub-optimal but still informative.

In our experiments, we compare a standard Diffusion Policy (DP) trained on FieldGen data against a reward-conditioned version (DP-R.). For the same collection time budget, the FieldGen-Reward pipeline generates 10 times the number of trajectories, with rewards uniformly distributed between 0 and 1. The DP-R. model is conditioned on this additional reward input during training.

As shown in Table\ref{tab:equal-time-reorg} and Figure\ref{fig:exp_reward}, the reward-conditioned model (DP-R.) consistently and significantly outperforms the standard DP model across all tasks and time budgets. For instance, after just 4 minutes of data collection, DP-R. achieves an average success rate of 79.2\%, a 29.2\% improvement over the standard DP. This performance gap widens with more data; at the 12-minute mark, DP-R. reaches a near-perfect average success rate of 95.9\%. These results demonstrate that explicitly modeling trajectory quality via rewards enables the policy to learn a more robust and effective representation of the task, leading to substantial gains in performance.
\section{CONCLUSIONS}

We propose FieldGen, a field-guided, semi-automated framework for real-world manipulation data generation. By decoupling reaching from fine manipulation and labeling reaching trajectories with cone-shaped (position) and spherical (orientation) attraction fields, FieldGen converts raw wrist-camera observations into high-quality $\langle$obs, action$\rangle$ pairs with minimal human effort. Real-world experiments show that FieldGen consistently outperforms teleoperation given equal collection time, scales more favorably with additional data, and generalizes better across start-pose, object-pose, and cross-instance settings. Ablations further indicate that cycloid-based trajectories improve success rates and motion smoothness over B'ezier alternatives. Compared with teleoperation, FieldGen acquires data more efficiently while imposing substantially lower operator effort. Finally, we validate that trajectories of varying quality generated by FieldGen-Reward improve model robustness and the model’s understanding of behavioral causality. FieldGen thus offers a practical path to faster collection and broader coverage for large-scale real-robot datasets. Future work will extend the fields to multi-step tasks, incorporate uncertainty-aware control, and explore scene synthesis (e.g., 3DGS/NeRF) for fully automated observation generation.



\bibliographystyle{plain}  
\bibliography{main}

\end{document}